\documentclass[runningheads]{llncs}
\usepackage{graphicx}
\usepackage{amsmath,amssymb} 
\usepackage[width=122mm,left=12mm,paperwidth=146mm,height=193mm,top=12mm,paperheight=217mm]{geometry}
\usepackage{color}

\usepackage{graphicx, amsmath, amssymb, caption, subcaption, multirow, overpic, textpos, booktabs,tabularx,colortbl}

\newlength\savewidth

\renewcommand{\paragraph}[1]{\vspace{1.25mm}\noindent\textbf{#1}}

\newcolumntype{x}[1]{>{\centering\arraybackslash}p{#1pt}}
\newcolumntype{y}[1]{>{\raggedright\arraybackslash}p{#1pt}}
\newcolumntype{z}[1]{>{\raggedleft\arraybackslash}p{#1pt}}

\newcommand{\app}{\raise.17ex\hbox{$\scriptstyle\sim$}}

\definecolor{deemph}{gray}{0.6}

\definecolor{baselinecolor}{gray}{.9}

\newcommand{\gr}{\rowcolor[gray]{.95}}

\begin{document}
\pagestyle{headings}
\mainmatter

\def\ACCV22SubNumber{79}  

\title{Towards Real-time High-Definition Image Snow
Removal: Efficient Pyramid Network with Asymmetrical Encoder-decoder Architecture} 
\titlerunning{ACCV-22 submission ID \ACCV22SubNumber}
\authorrunning{ACCV-22 submission ID \ACCV22SubNumber}


\author{Tian Ye\inst{1}$^\dag$ \and
Sixiang Chen$^\dag$\inst{1} \and
Yun Liu\inst{2}\and
Yi Ye\inst{1}\and \\
Erkang Chen\thanks{Corresponding author.$^\dag$Equal contribution.} \inst{1}
}

\authorrunning{Tian Ye et al.}
\titlerunning{Abbreviated paper title}
%
\institute{$^1$School of Ocean Information Engineering,
    Jimei University, Xiamen, China\\
    $^2$ Southwest University School of Artificial Intelligence, Chonqing, China \\}
    
\maketitle

\begin{abstract}
In winter scenes, the degradation of images taken under snow can be pretty complex, where the spatial distribution of snowy degradation is varied from image to image. Recent methods adopt deep neural networks to directly recover clean scenes from snowy images. However, due to the paradox caused by the variation of complex snowy degradation, achieving reliable High-Definition image desnowing performance in real time is a considerable challenge. We develop a novel Efficient Pyramid Network with asymmetrical encoder-decoder architecture for real-time HD image desnowing. The general idea of our proposed network is to utilize the multi-scale feature flow fully and implicitly mine clean cues from features. Compared with previous state-of-the-art desnowing methods, our approach achieves a better complexity-performance trade-off and effectively handles the processing difficulties of HD and Ultra-HD images.

The extensive experiments on three large-scale image desnowing datasets demonstrate that our method surpasses all state-of-the-art approaches by a large margin both quantitatively and qualitatively, boosting the PSNR metric from 31.76 dB to 34.10 dB on the CSD test dataset and from 28.29 dB to 30.87 dB on the SRRS test dataset.
\end{abstract}

\section{Introduction}

In nasty weather scenes, snow is an essential factor that causes noticeable visual quality degradation. Degraded images captured under snow scenes significantly affect the performance of high-level computer vision tasks~\cite{ouyang2013joint,redmon2016you,shafiee2017fast,huang2020nms,lan2020global,szegedy2013deep}.

Snowy images suffer more complex degradation by various factors than common weather degradation,i.e., haze and rain. According to previous works, snow scenes usually contain the snowflake, snow streak, and veiling effect. The formation of snow can be modeled as:
\begin{equation}
\label{eq1}
\mathbf{I}(x)=\mathbf{K}(x) \mathbf{T}(x)+\mathbf{A}(x)(1-\mathbf{T}(x)),
\end{equation}
where $\mathbf{K}(x)=\mathbf{J}(x)(1-\mathbf{Z}(x) \mathbf{R}(x))+\mathbf{C}(x) \mathbf{Z}(x) \mathbf{R}(x)$, $\mathbf{I}(x)$ denotes the snow image, $\mathbf{K}(x)$ denotes the veiling-free snowy image,, $\mathbf{A}(x)$ is the atmospheric light, and $\mathbf{J}(x)$ is the scene radiance. $\mathbf{T(x)}$ is the transmission map. $\mathbf{C}(x)$ and $\mathbf{Z}(x)$ are the chromatic aberration map for snow images and the snow mask, respectively. $\mathbf{R}(x)$ is the binary mask, presenting the snow location information. 

As described in Eq.\ref{eq1}, the chromatic aberration degradation of snow and the veiling effect of haze are mixed in an entangled way.  Existing snow removal methods can be categorized into two classes: model-based methods and model-free methods. For model-based methods~\cite{zheng2013single,liu2018desnownet,chen2020jstasr}, JSTASR~\cite{chen2020jstasr} tries to recover a clean one from a snow image in an uncoupled way. Utilizing the veiling effect recovery branch to recover the veiling effect-free image, and the snow removal branch to recover the snow-free image. However, the divide and conquer strategy ignores the influence of inner entanglement degradation in snow scenes, and complicated networks by hand-craft design results in unsatisfactory model complexity and inference efficiency. For model-free methods~\cite{li2020all,chen2021all,valanarasu2021transweather}, HDCW-Net~\cite{chen2021all} proposed the hierarchical decomposition paradigm, which leverages frequency domain operations to extract clean features for better understanding the various diversity of snow particles. But the dual-tree complex wavelet limits the inference performance of HDCW-Net and it still has scope for improvement in its performance.

The single image snow removal methods have made remarkable progress recently. Yet, there are few studies on the efficient single image snow removal network, which attract us to explore the following exciting topic:
\\[4pt]
\textit{How to design an \textbf{efficient} network to \textbf{effectively} perform single image desnowing?}
\\[4pt]
Most previous learning-based methods~\cite{chen2021all,chen2020jstasr,valanarasu2021transweather} hardly achieve real-time inference efficiency with HD resolution, even running on the expensive advanced graphics processing unit. We present a detailed run-time comparison in the experiment section to verify this point. Most methods can not perform real-time processing ability to handle High-Definition degraded images. In this manuscript, we propose an efficient manner to perform \textbf{HD(1280$\times$720)} resolution image snow removal in real time, which is faster and more deployment-friendly than previous methods. Moreover, our method is the first desnowing network which can handle \textbf{$UHD(4096\times2160)$} image processing problem.

Previous desnowing networks~\cite{chen2021all,chen2020jstasr} usually only have 2 to 3 scale-level, which limits the desnowing performance and inferencing efficiency. Different from the previous mainstream design of desnowing CNN architecture, the proposed network owns five scale levels, which means the smallest feature resolution of the feature flow in our pyramid network is only $\frac{1}{32}$ of input snow image. Sufficient multi-scale information brings impressive representation ability. Furthermore, the efficient and excellent basic block is also a significant factor for network performance and actual efficiency. Thus we propose a Channel Information Mining Block (\textbf{CIMB}) as our basic block to mine clean cues from channel-expanded features, which is inspired by NAFNet~\cite{chen2022simple}. Besides that, we propose a novel External Attention Module (EAM) to introduce reliable information from original degraded images to optimize feature flows in the pyramid architecture. The proposed EAM can adaptively learn more useful sample-wise features and emphasize the most informative region on the feature map for image desnowing. 

Motivated by the proposed efficient and effective components, our method achieves excellent desnowing performance. Compared with the previous best method transweather~\cite{valanarasu2021transweather}, the proposed method has better quantitative results (\textbf{30.61dB/0.91 vs. 35.60dB/0.95}). And as shown in Figure.~\ref{fig:psnr_compelxity_trade_off}, compared with previous state-of-the-art desnowing methods, our method achieves a better complexity-performance trade-off .

The main contributions of this paper are summarized as follows:

\begin{figure}[t!]
    \centering
    \includegraphics[width=11cm]{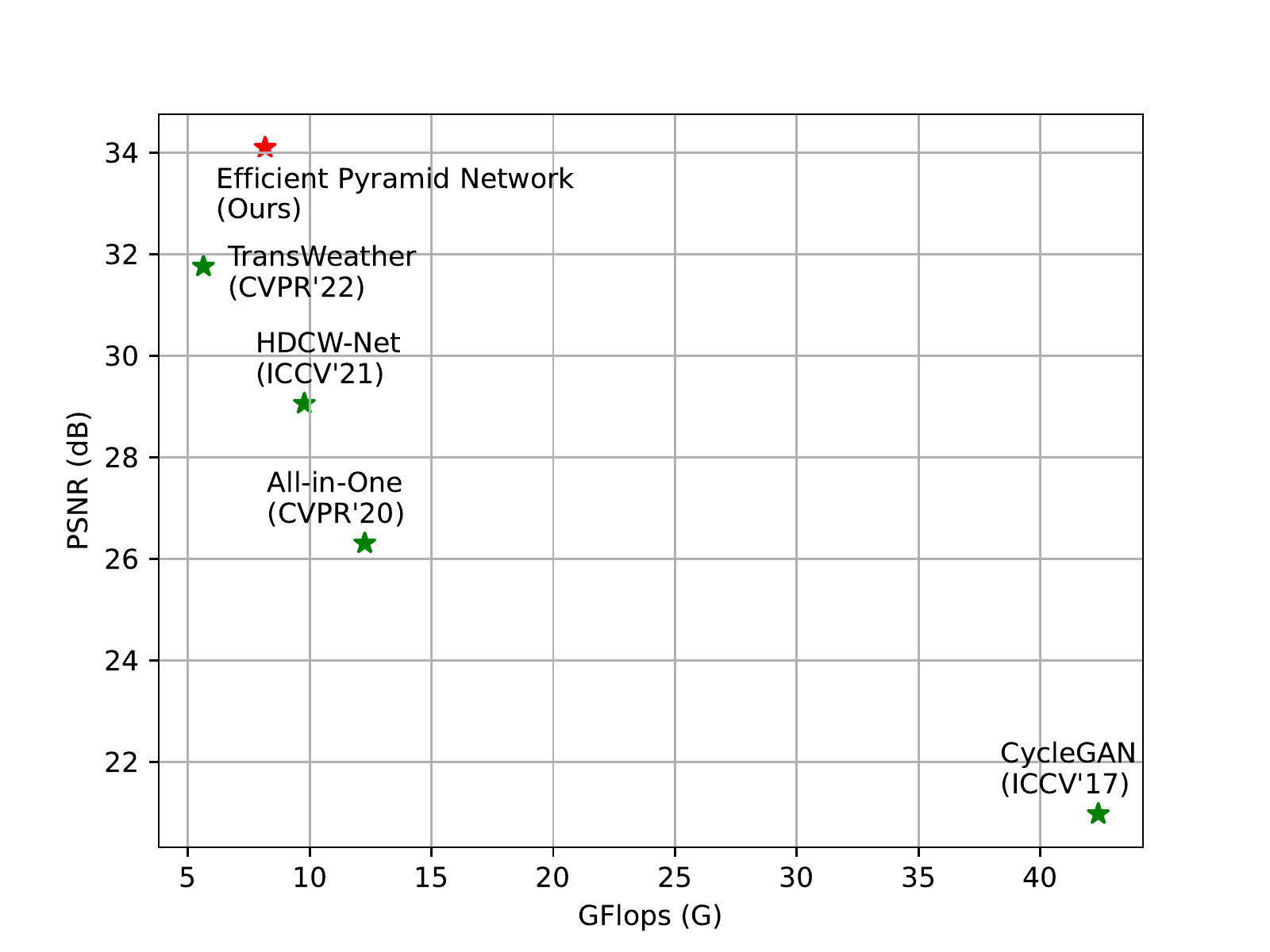}
    \caption{Trade-off between performance vs: number of operations on CSD (2000)~\cite{chen2021all} testing dataset. Multi-adds are
calculated on $256\times256$ image. The results show the superiority of
our model among existing methods}
    \label{fig:psnr_compelxity_trade_off}
\end{figure}

\begin{itemize}

\item a) We propose a Channel Information Mining Block (CIMB) explore clean cues efficiently. Furthermore, the External Attention Module (EAM) is proposed to introduce external instructive information to optimize features. Our ablation study demonstrates the effectiveness of CIMB and EAM.
    
\item b) We propose the Efficient Pyramid Network with asymmetrical encoder-decoder architecture, which achieves real-time High-Definition image desnowing.

\item c) Our method achieves the best quantitative results compared with the state-of-the single image desnowing approaches.
    
\end{itemize}

\section{Related Works}
Traditional methods usually make assumptions and use typical priors to handle the ill-posed nature of the desnowing problem. One of the limitations of these prior-based methods is that these methods can not hold well for natural scenes containing various snowy degradation and hazy effect.
Recently, due to the impressive success of deep learning in computer vision tasks, many learning-based approaches have also been proposed for image desnowing~\cite{liu2018desnownet,chen2020jstasr,chen2021all,li2020all}. In these methods, the key idea is to directly learn an effective mapping between the snow image and the corresponding clean image using a solid CNN architecture. However, these methods usually involve complex architecture and large-kernel size of essential convolution components and consume long inferencing times, which cannot cover the deployment demand for real-time snow image processing, especially for High-Definition (HD) and Ultra-High-Definition (UHD) images.

The first desnowing network is named DesnowNet~\cite{liu2018desnownet}, which focuses on removing translucent and snow particles in sequence based on a multi-stage CNN architecture. Li \textit{et al.} propose an all-in-one framework to handle multiple degradations, including snow, rain, and haze. JSTASR~\cite{chen2020jstasr} propose a novel snowy scene imaging model and develop a size-aware snow removal network that can handle the veiling effect and various snow particles. HDCW-Net~\cite{chen2021all} performs single image snow removal by utilizing the dual-tree wavelet transform and designing a multi-scale architecture to handle the various degradation of snow particles. Most previous methods are model-based, which limits the representation ability of CNNs. Moreover, the wavelet-based method is not deployment-friendly for applications. 

\section{Proposed Method}
This section will first introduce the Channel Information Mining Block (CIMB) and External Attention Module. Then, we present the proposed pyramid architecture. Worth noting that profit from the asymmetrical encoder-decoder design and efficient encoder blocks, the proposed method is faster and more effective than previous CNN methods.

\subsection{Channel Information Mining Block}

\begin{figure}
    \centering
    \includegraphics[width=\textwidth]{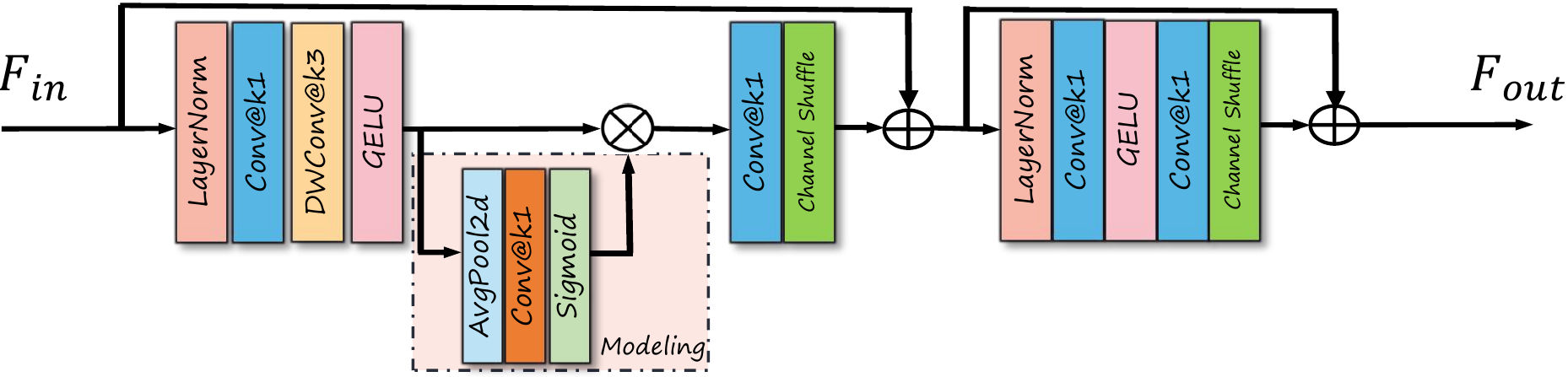}
    \caption{The proposed Channel Information Mining Block (\textbf{CIMB}). The \textit{DWConv} denotes the depth-wise convolution operation, and the kernel size of the convolution is denoted by $@k*$. The design of our CIMB is motivated by NAFNet~\cite{chen2022simple}.}
    \label{fig:CIMB}
\end{figure}

Previous classical basic block in desnowing networks~\cite{liu2018desnownet,chen2020jstasr,chen2021all} usually utilize large kernel-size convolution and frequency-domain operations. In contrast with the previous design, the proposed Channel Information Mining Block (CIMB) focus on how to effectively mine clean cues from incoming features with minimum computational cost.

As shown in Fig.~\ref{fig:CIMB}, let's denote the input feature as $F_{in}^{c}$ and output feature as $F_{out}^c$. The computational process of CIMB can be presented by:
\begin{equation}
    F_{out} = CIMB (F_{in}),
\end{equation}
Our design is simple and easy to implement in a widely-used deep learning framework. We utilize Layer Normalization to stable the training of the network and use the convolution with the kernel size of $1 \times 1$ to expand the channel of feature maps from $c$ to $\alpha c$, where the $\alpha$ is the channel-expand factor, is set as 2 in our all experiments. 
The channel-expanding way is crucial to motivate state-of-the-art performance because we found that high-dimension information mining is significantly suitable for single image desnowing. We further utilize channel information modeling to model the distribution of snow degradation and explore clean cues in the channel dimension. The channel-expanding design is motivated by NAFNet~\cite{chen2022simple}, which is an impressive image restoration work. Nevertheless, it ignores achieving efficient information interaction across different channels. We deeply realize the lack of effective channel interaction and introduce channel shuffle operation to achieve efficient and effective channel information interaction. Our ablation study demonstrates that the clever combination of channel-expanding and channel-shuffle achieves better performance with little computational cost by channel-shuffle.  

\subsection{External Attention Module}
\begin{figure}
    \centering
    \includegraphics[width=\textwidth]{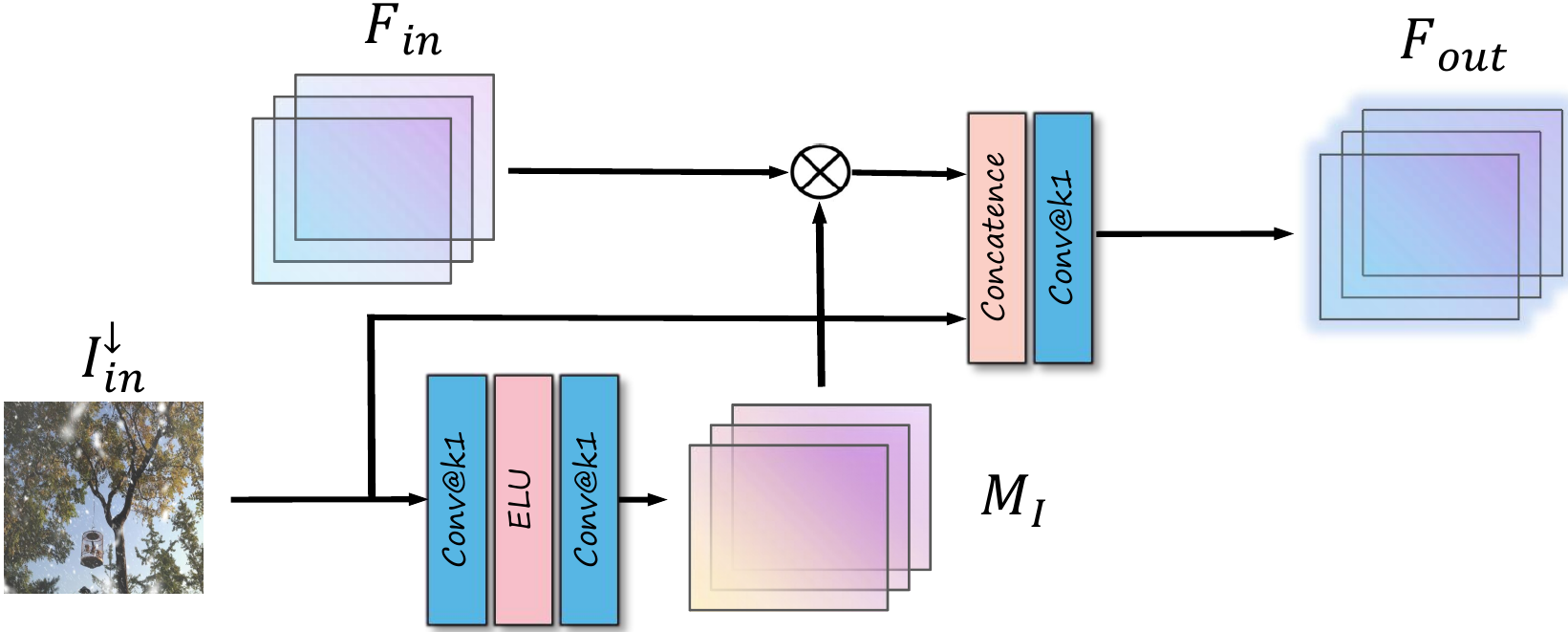}
    \caption{The proposed External Attention Module (\textbf{EAM}). The kernel size of the convolution is denoted by $@k*$.The tensor size of the external attention map $M_{I}$ is the same as the incoming feature $F_{in}$ and generated feature $F_{out}$. }
    \label{fig:EAM}
\end{figure}
In the past few years, more and more attention mechanism~\cite{hu2018squeeze,woo2018cbam,woo2018cbam} have been proposed to improve the representation ability of the convolution neural network. However, most attention module~\cite{hu2018squeeze,woo2018cbam,hou2021coordinate,ffa-net} only focus on exploring useful information from incoming features, which ignores capturing and utilizing implicit instruction information from original input images. Moreover, information loss in multi-scale architecture cannot be avoided; thus, we utilize down-sampled snow degraded images to introduce original information, relieving the information loss by multi-scale feature flow design.

As shown in Fig.~\ref{fig:EAM}, the External Attention Module is a plug-and-play architectural unit that can be directly used in CNNs for image restoration tasks. Specifically, we perform a series of operations on the down-sampled $I_{i n}^{\downarrow}$ to generate the external attention map $\mathcal{M}^{H \times W \times C}_{I}$:
\begin{equation}
\text { Conv } \circ \mathrm{ELU} \circ \text { Conv with } 1 \times 1: I_{i n}^{\downarrow} \rightarrow \mathcal{M}^{H \times W \times C}_{I},
\end{equation}
where the $I_{i n}^{\downarrow}$ is the down-sampled original image, whose spatial size is the same as the incoming feature map $F_{in}$. Then we multiply
the $\mathcal{M}^{H \times W \times C}_{I}$ with the $F_{in}$:

\begin{equation}
{F}_{\text {att}}^{H \times W \times C}=
\mathcal{M}^{H \times W \times C}_{I} \cdot 
{F}_{\text {in }}^{H \times W \times C},
\end{equation}

where the ${F}_{\text {att}}^{H \times W \times C}$ is the scaled feature map. And we further utilize $I_{i n}^{\downarrow}$ and channel-wise compression to introduce the original information:
\begin{equation}
\text { Concatence: } {I}_{in}^{\downarrow}+{F}_{att}^{H \times W \times C} \rightarrow {F}_{fusion}^{H \times W \times (C+3)}. \end{equation}
And a convolution with kernel size of $3\times3$ is be used to compress the dimension of $F_{fusion}$ and get the final output feature $F_{out}$:
\begin{equation}
\text { Conv:}    
{F}_{fusion}^{H \times W \times (C+3)} \rightarrow {F}_{out}^{H \times W \times C}.
\end{equation}
The benefits of our EAM are twofold:
i) Fully utilize original degradation information to instruct the feature rebuilding explicitly. 
ii) Relief the information loss by repeat down-up sampling in multi-scale architecture.
Please refer to our experiments section for the ablation study about the External Attention Module, which demonstrates the effectiveness of our proposed EAM.

\subsection{Efficient Pyramid Network with Asymmetrical Encoder-decoder Architecture}

\begin{figure}[t!]
    \centering
    \includegraphics[width=\textwidth]{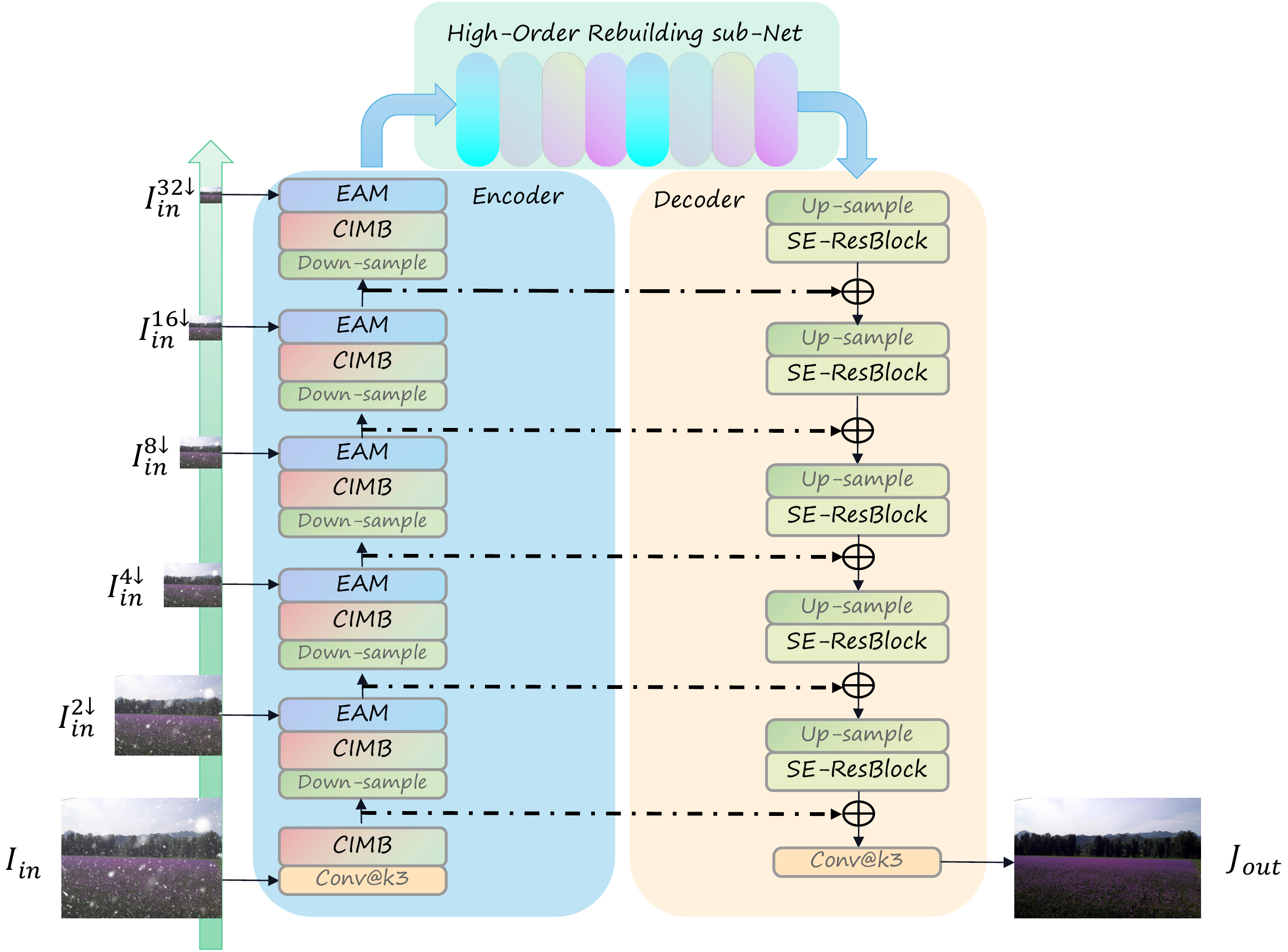}
    \caption{Overview of the proposed Efficient Pyramid Network. }
    \label{fig:Architecture}
\end{figure}

\paragraph{Efficient Pyramid Network.}
U-Net style architectures can bring sufficient multi-scale information compared with single-scale-level architectures. However, the fully symmetrical architecture of U-Net style architectures results in redundant computational costs. Unlike previous mainstream designs~\cite{chen2022simple,wu2021contrastive} in the image restoration area, we develop an efficient pyramid architecture to explore clean cues from features of 5 scale levels, which is a more efficient way to handle the complex degradation, aka, snowy particles, and uneven hazy effect. 

As shown in Fig.~\ref{fig:Architecture}, our efficient pyramid network consists of three parts: encoder, decoder, and High-Order Rebuilding Sub-Net. Every scale-level encoder block only has a CIMB and an EAM in the encoder stage. For the decoder, we utilize ResBlock with SE attention~\cite{hu2018squeeze} as our basic decoder block. Our decoder is fast and light. In the following sections, we further present the straight idea about Asymmetrical E-D design and High-Order Rebuilding Sub-Net.

\paragraph{Asymmetrical Encoder-decoder Architecture.}
The symmetrical encoder-decoder architecture has been proven work well in many CNNs~\cite{wu2021contrastive,chen2022simple,chen2021all}. Most methods tend to add more conv-based blocks in every scale level with a symmetrical design to improve the model performance further. However, directly utilizing symmetrical architecture is not the best choice for our aims for the following reasons. First, we aim to process \textit{\textbf{HD ($1280\times720$)}} snowy images in real time, so we have to make trade-offs between performance and model complexity. Second, we found that a heavy encoder with a light decoder has a better representation ability to explore clean cues than a symmetrical structure. Thanks to the asymmetrical ED architecture, our Efficient Pyramid Network performs well on HD snow images in real time.

\paragraph{High-Order Rebuilding Sub-Net.}
It is critical for an image restoration network to exploit clean cues from the latent features effectively. Here, we design an effective High-Order Rebuilding Sub-Net to rebuild clean features in high-dimension space further. Our High-Order Rebuilding Sub-Net comprises 20 Channel Information Mining Block of 512 dimensions. Due to latent features only having $\frac{1}{32}\times$ resolution size than original input images, our High-Order Rebuilding Sub-Net quickly performs clean cues mining.

\section{Loss Function}
We only utilize the Charbonnier loss~\cite{charbonnier1994two} as our basic reconstruction loss:
\begin{equation}
\mathcal{L}_{\text {char }}=\frac{1}{N} \sum_{i=1}^{N} \sqrt{\left\|J_{out}^{i}-J_{gt}^{i}\right\|^{2}+\epsilon^{2}},
\end{equation}
with constant $\epsilon$ empirically set to $1e^{-3}$ for all experiments. And $J_{gt}^i$ denotes the ground-truth of $J_{out}^i$ correspondingly.

\section{Experiments}
\subsection{Datasets and Evaluation Criteria}
We choose the widely used PSNR and SSIM as experimental metrics to measure the performance of our network. We train and test the proposed network on three large datasets: CSD~\cite{chen2021all}, SRRS~\cite{chen2020jstasr}
 and Snow100k~\cite{chen2020jstasr}, following the benchmark-setting of latest desnowing methods~\cite{chen2021all} for authoritative evaluation. Moreover, we re-train the latest bad weather removal method TransWeather (CVPR'22)~\cite{valanarasu2021transweather} to make a better comparison and analysis. Worth noting that we reproduce the TransWeather-based for a fair comparison, provided by the official repository of TransWeather~\cite{valanarasu2021transweather}. 

\subsection{Implementation Details}
We augment the training dataset by randomly rotating by 90,180,270 degrees and horizontal flip. The training patches with the size $256\times256$ are extracted as input paired data of our network. We utilize the AdamW optimizer with an initial learning rate of $4\times10^{-4}$ and adopt the CyclicLR to adjust the learning rate progressively, where on the mode of triangular, the value of gamma is $1.0$, base momentum is 0.9, the max learning rate is $6\times10^{-4}$ and base learning rate is the same as the initial learning rate. We utilize the Pytorch framework to implement our network with 4 RTX 3080 GPU with a total batch size of 60. For channel settings, we set the channel as $[16,32,64,128,256,512]$ in each scale-level stage respectively. 

\subsection{Performance Comparison}
In this section, we compare our Efficient Pyramid Network method with the state-of-the-art image desnowing methods of ~\cite{liu2018desnownet,chen2020jstasr,chen2021all}, classical image translation method of ~\cite{engin2018cycle} and the bad weather removal methods of ~\cite{li2020all,valanarasu2021transweather}.

\subsubsection{Visual Comparison with SOTA methods}
\begin{figure}
    \centering
    \includegraphics[width=\textwidth]{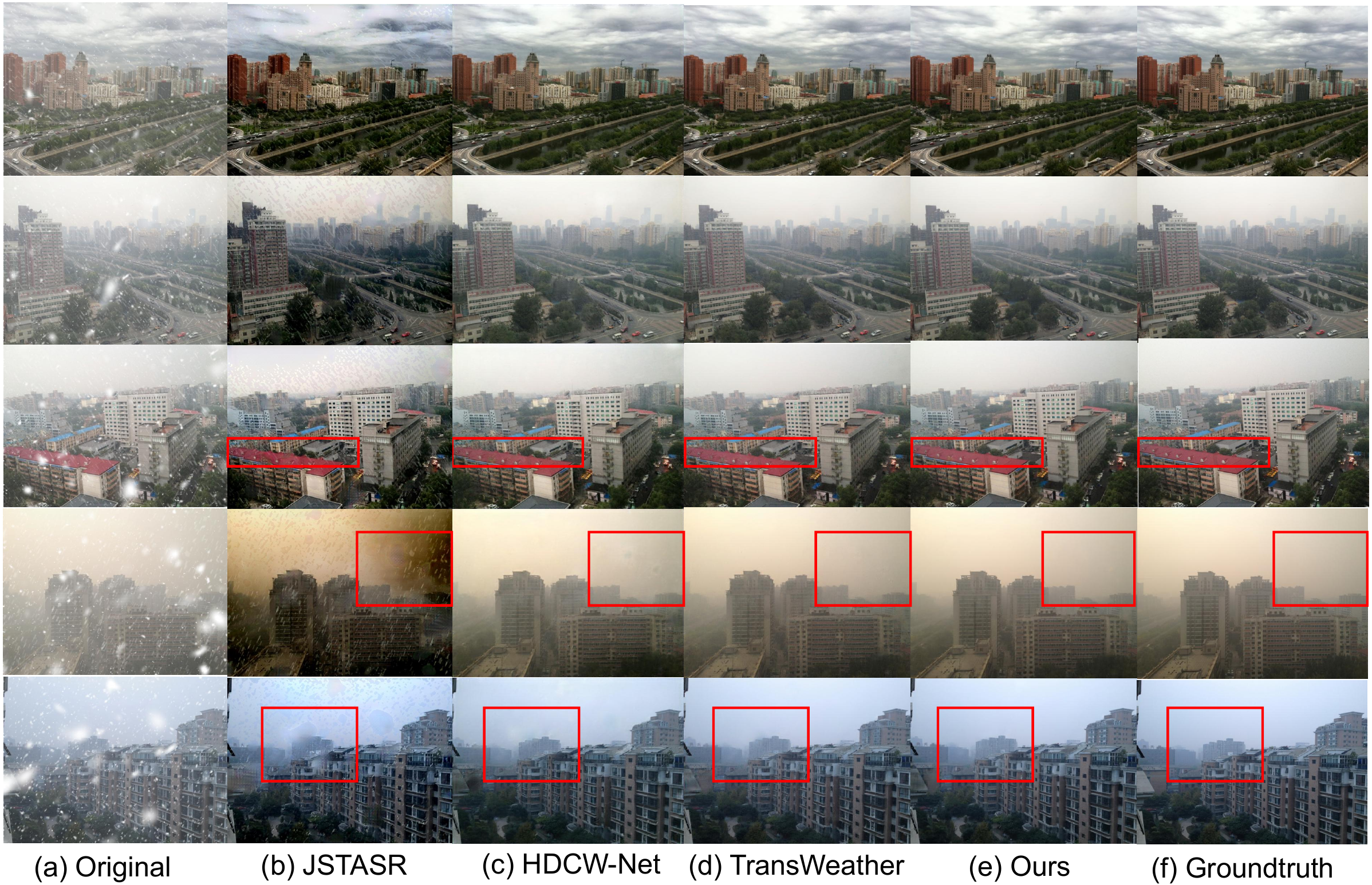}
    \caption{Visual comparisons on results of various methods (b-e) and our proposed network(e) on synthetic winter photos. Please zoom in for a better
illustration.}
    \label{fig:SynComparison}
\end{figure}

We compare our method with the state-of-the-art image desnowing methods on the quality of restored images, presented in Fig.~\ref{fig:SynComparison} and ~\ref{fig:SynComparison}. Our method generates the most natural desnowing images compared to other methods. The proposed method effectively restores the degraded area of the snow streaks or snow particles in both synthetic and authentic snow images.

\subsubsection{Quantitative Results Comparison}
In Table ~\ref{tab:ComparisonDesnowing}, we summarize the performance of out Efficient Pyramid Network and SOTA methods on CSD~\cite{chen2021all}, SRRS~\cite{chen2020jstasr} and Snow 100K~\cite{liu2018desnownet}. Our method achieves the best performance with 34.10dB PSNR and 0.95 SSIM on the test dataset of CSD. Moreover, it achieves the best performance with 30.87dB PSNR, 33.62 dB PSNR, and 0.94 SSIM, 0.96 SSIM on SRRS and Snow100k test datasets.

\begin{table*}[t!]
\caption{{Quantitative comparisons of our method with the state-of-the-art desnowing methods on CSD,SRRS and Snow 100K desnowing dataset (PSNR(dB)/SSIM). The best results are shown in \textbf{bold}, and second best results are \underline{underlined}.}}
		\centering
		 \resizebox{12cm}{!}{
\begin{tabular}{l|cc|cc|cc|c|c}
\toprule[1.2pt]

\multicolumn{1}{c|}{\multirow{2}{*}{Method}} & \multicolumn{2}{c|}{CSD(2000)} & \multicolumn{2}{c|}{SRRS (2000)} & \multicolumn{2}{c|}{Snow 100K (2000)} & \multirow{2}{*}{\#Param} & \multirow{2}{*}{\#GMacs}\\ \cline{2-7}
\multicolumn{1}{c|}{}                        & PSNR           & SSIM          & PSNR            & SSIM           & PSNR               & SSIM             &                          \\ \hline
(TIP'18)Desnow-Net~\cite{liu2018desnownet}                                   & 20.13          & 0.81         & 20.38           & 0.84           & 30.50              & 0.94             & 26.15 M           &1717.04G          \\
(ICCV'17)CycleGAN~\cite{engin2018cycle}                                     & 20.98          & 0.80          & 20.21           & 0.74           & 26.81              & 0.89             & 7.84M        &42.38G     \\

(CVPR'20)All-in-One~\cite{li2020all}                                   & 26.31          & 0.87          & 24.98           & 0.88           & 26.07              & 0.88             & 44 M             &12.26G           \\
(ECCV'20)JSTASR~\cite{chen2020jstasr}                                       & 27.96          & 0.88          & 25.82           & 0.89           & 23.12              & 0.86             & 65M           &-         \\
(ICCV'21)HDCW-Net~\cite{chen2021all}                                     &{29.06}    &{0.91}    &{27.78}     &\underline{0.92}     &{31.54}        &{\underline{0.95}}       & 699k     &9.78G   \\ 

(CVPR'22)TransWeather                                     & \underline{31.76}    & \underline{0.93}    &{\underline{28.29}}     & \underline{0.92}     &{\underline{31.82}}        &{0.93}       &21.9M     &5.64G   \\ \hline

\gr Ours                              & {\textbf{34.10}}          & {\textbf{0.95}}          & {\textbf{30.87}}               &{\textbf{0.94}}              & {\textbf{33.62}}                  &{\textbf{0.96}}                & 66.54M    & 8.17G                 \\ 

\bottomrule
\end{tabular}
	}
\label{tab:ComparisonDesnowing}
\end{table*}

\subsubsection{Run-time Discussion}
In Table ~\ref{tab:run_time_comparison1},~\ref{tab:run_time_comparison2},~\ref{tab:run_time_comparison3}, we present detailed run-time and model-complexity comparison with different processing resolution settings. Worth noting that TransWeather~\cite{valanarasu2021transweather} can not handle \textbf{UHD($4096\times2160$)} degraded images, as shown in Table ~\ref{tab:run_time_comparison2}, although it is minorly faster than ours when input image is HD or smaller size. As shown in Table ~\ref{tab:run_time_comparison1}, our Efficient Pyramid Network achieve real-time performance when processing HD resolution images.

\begin{table}[h!]

\caption{\scriptsize{\textbf{Comparison of Inference time, GMACs (fixed-point multiply accumulate operations performed persecond) and Parameters when input HD ($1280\times 720$) images}. Our method achieves the best runtime-performance trade-off compared to the state-of-the-art approaches. The time reported in the table corresponds to the time taken by each model feed forward an image of dimension $1280\times720$ during the inference stage. We perform all inference testing on an A100 GPU for a fair comparison. 
Notably, we utilize the \textit{torch.cuda.synchronize()} API function to get accurate feed forward run-time.
}}
\centering
\label{tab:run_time_comparison1}
\vspace{0.3cm}
{
    \begin{tabular}{c|c|c|c}
    \hline
    Method  & Inf. Time(in s) & GMACs(G) & Params(M) \\ \hline
    TransWeather~\cite{valanarasu2021transweather} & 0.0300           & 79.66     & 21.9      \\
    Ours    & 0.0384       & 113.72     & 65.56      \\ \hline
    \end{tabular}
}

\end{table}

\begin{table}[h!]

\caption{\scriptsize{\textbf{Comparison of Inference time, GMACs (fixed-point multiply accumulate operations performed persecond) and Parameters when input UHD ($4096\times2160$) images}. We perform all inference testing on an A100 GPU for a fair comparison.}}
\centering
\label{tab:run_time_comparison2}
\vspace{0.3cm}
{
    \begin{tabular}{c|c|c|c}
    \hline
    Method  & Inf. Time(in s) & GMACs(G) & Params(M) \\ \hline
    TransWeather~\cite{valanarasu2021transweather} & Out of Memory           & -     & 21.9      \\
    Ours    & 0.23       & 1097.03     & 65.56      \\ \hline
    \end{tabular}
}

\end{table}

\begin{table}[h!]

\caption{\scriptsize{\textbf{Comparison of Inference time, FPS and Parameters when input small ($512\times672$) images}. Following the testing platform 
of HDCW-Net~\cite{chen2021all}, we perform all inference testing on a RTX 1080ti GPU for a fair comparison.}}
\centering
\label{tab:run_time_comparison3}
\vspace{0.3cm}
{
    \begin{tabular}{c|c|c|c}
    \hline
    Method  & Inf. Time(in s) & FPS & Params(M) \\ \hline
    JSTASR~\cite{chen2020jstasr} &0.87           & 1.14    & 65  \\
    HDCW-Net~\cite{chen2021all} &0.14           &7.14     &0.699   \\
    Ours    & 0.0584       & 17.11     & 65.56      \\ \hline
    \end{tabular}
}

\end{table}

\begin{figure}
    \centering
    \includegraphics[width=\textwidth]{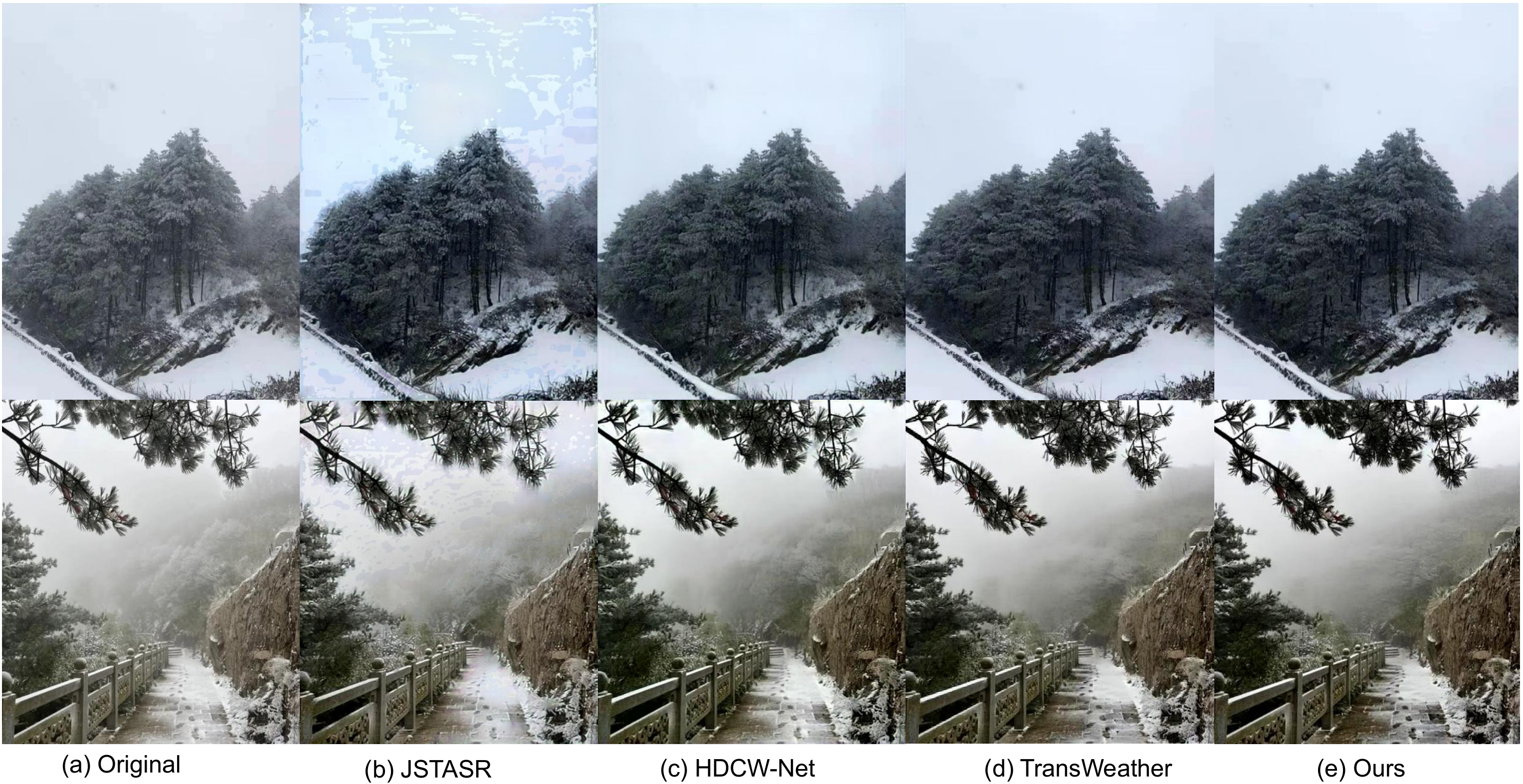}
    \caption{Visual comparisons on results of various methods (b-d) and our method(e) on real winter photos. Please zoom in for a better
illustration.}
    \label{fig:RealComparison}
\end{figure}

\begin{figure}[h!]
    \centering
    \includegraphics[width=\textwidth]{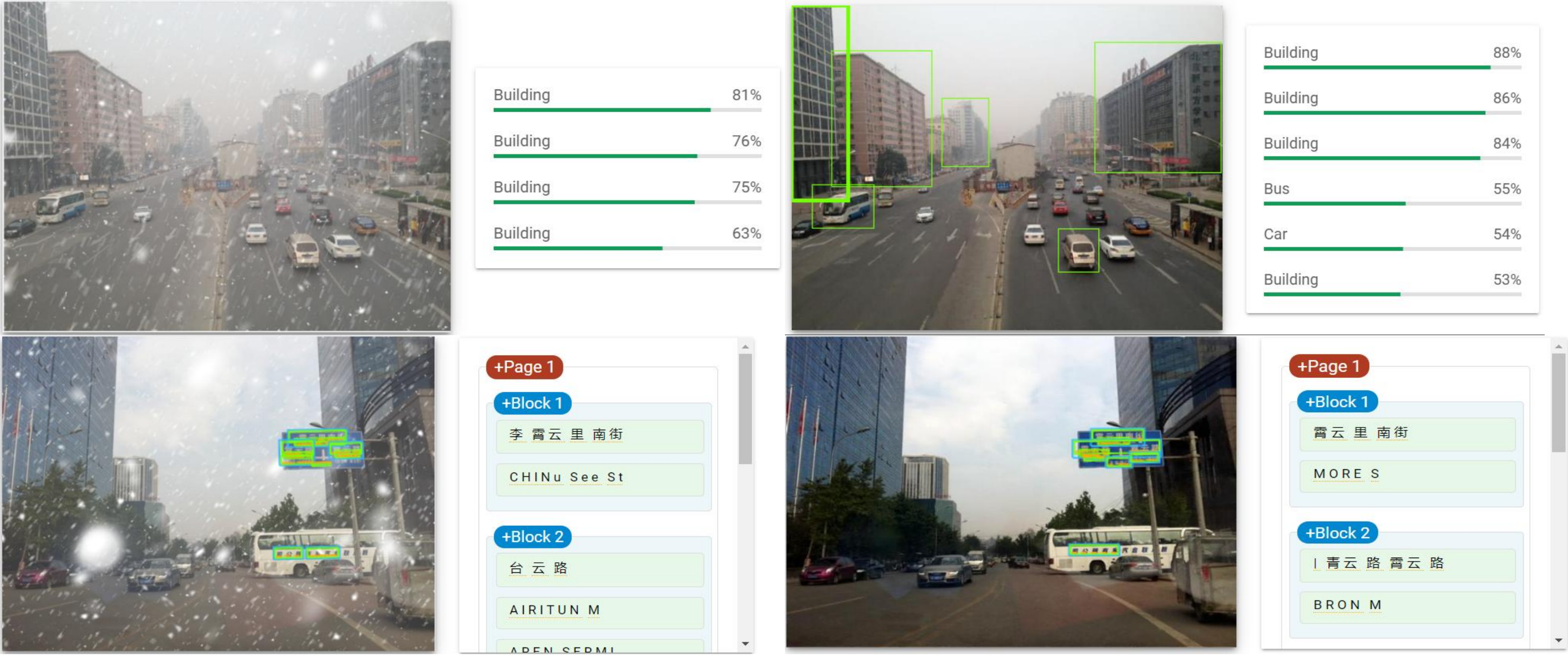}
    \caption{Synthetic snow images (left)
and corresponding desnowing results (right) by the proposed method with the high-level vision task results and confidences supported by Google Vision API.}
    \label{fig:highleveltasks}
\end{figure}

\subsubsection{Model Complexity and Parameter Discussion}
For real-time deployment of CNNs, computational complexity is
an important aspect to consider. To check the computational complexity, we tabulate the GMAC in Table~\ref{tab:ComparisonDesnowing} for previous state-of-the-art methods and our proposed method when the input image is $256 \times 256$. We note that our network has less computational complexity when compared to previous methods. Notably, Desnow-Net is highly computationally complex than ours, even though it has less number of parameters.

\subsection{Quantifiable Performance for High-level Vision Tasks} As shown in Fig.~\ref{fig:highleveltasks}, we show subjective but quantifiable results for a concrete demonstration, in which the vision task results and corresponding confidences are both supported by Google Vision API. Our comparison illustrates that these snowy degradations could impede the actual performance of high-level vision tasks. And our method could significantly boost the performance of high-level vision tasks. 

\subsection{Ablation Study}
For a reliable ablation study, we utilize the latest desnowing dataset,\textit{i,e.}, CSD~\cite{chen2021all} dataset as the benchmark
for training and testing in all ablation study experiments.
\begin{table}[h!]

\setlength{\abovecaptionskip}{0.0cm} 
\setlength{\belowcaptionskip}{0cm}
\caption{{Configurations of the proposed Channel Information Mining Block.}}
\centering
\resizebox{12cm}{!}{
\begin{tabular}{cccccc}
\toprule
\textbf{Metric}    & \textbf{wo LN} & \textbf{GELU$\rightarrow$ReLU} &  \textbf{wo Channel Shuffle}& \textbf{Ours} \\ \hline\hline
\textbf{CSD(2000) PSNR/SSIM} & \textbf{33.79/0.93}   & \textbf{33.76/0.94}   & \textbf{32.61/0.93}      & \textbf{34.10 /0.95}      \\ 
\bottomrule
\end{tabular}}
\label{tab:abalation1}
\end{table}

\subsubsection{Configurations of the Channel Information Mining Block} In Table~\ref{tab:abalation1}, we present the quantitative results of different configuration settings about CIMB. Specifically, we remove the Layer Normalization (\textbf{wo LN}), replace the GELU with ReLU (\textbf{GELU $\rightarrow$ ReLU}) and remove the channel shuffle operation(\textbf{wo Channel Shuffle}). 
From results of Table.~\ref{tab:abalation1}, we can find that using Channel Shuffle operation brings 1.49 dB PSNR improvement.

\begin{table}[h!]
\setlength{\abovecaptionskip}{0.0cm} 
\setlength{\belowcaptionskip}{0cm}
\caption{{Verification for the proposed External Attention Module.}}
\centering
\resizebox{12cm}{!}{
\begin{tabular}{cccccc}
\toprule
\textbf{Metric}    
& \textbf{wo Concat $I_{in}^{\downarrow}$} 
& \textbf{$I_{in}^{\downarrow} \rightarrow F_{in}$} 
&  \textbf{ELU $\rightarrow$ ReLU} & \textbf{Ours} \\ \hline\hline
\textbf{CSD(2000) PSNR/SSIM} & \textbf{32.16 /0.93}   & \textbf{32.89 /0.94}   & \textbf{33.96 /0.94}      & \textbf{34.10 /0.95}      \\ 

\bottomrule
\end{tabular}}
\label{tab:abalation2}
\end{table}

\subsubsection{Verification of key designs for the proposed External Attention Module.} In Table ~\ref{tab:abalation2}, we present our verification about designs of the External Attention Module. (a)\textbf{wo Concat $I_{in}^{\downarrow}$}. To verify the effectiveness of introducing down-sampled original images, we remove the setting of image-feature fusion by concatence. (b)\textbf{$I_{in}^{\downarrow} \rightarrow F_{in}$}. To verify the key idea that generate a external attention map from the original degraded image to optimize current scale features, we replace the $I_{in}$ with incoming $F_{in}$. (c)  \textbf{ELU $\rightarrow$ ReLU}. We replace the $ELU$ with $ReLU$ to explore the influence of non-liner function for network performance.

\begin{table}[h!]
\caption{{Ablation study of the High-Order Rebuilding Sub-Net.}}
\centering
\resizebox{11cm}{!}{
\begin{tabular}{cccccc}
\toprule
\textbf{Block Num.}    & \textbf{0} & \textbf{6} &  \textbf{12}& \textbf{Ours} \\ \hline\hline
\textbf{CSD(2000) PSNR/SSIM} & \textbf{29.76 /0.93}   & \textbf{32.41/0.93}   & \textbf{33.69/0.94}      & \textbf{34.10/0.95}      \\ 

\textbf{Params.(M)} & \textbf{12.81}   & \textbf{28.64}   & \textbf{44.46}      & \textbf{65.56}      \\

\bottomrule
\end{tabular}}
\label{tab:abalation3}
\end{table}

\subsubsection{Ablation study of the High-Order Rebuilding Sub-Net.} In Table~\ref{tab:abalation3}, we explore the influence of the depth of High-Oder Rebuilding Sub-Net. We found that deeper architecture has better performance on image desnowing. Due to 5 level down-sampling design, simply stacking more blocks cannot result in an unacceptable computational burden.

\begin{table}[h!]
\caption{{Ablation study of the Efficient Pyramid Network.}}
\centering
\resizebox{12cm}{!}{
\begin{tabular}{cccccc}
\toprule
\textbf{Metric}    & \textbf{CIMB $\rightarrow$ SE-ResBlock} & \textbf{wo EAM} &  \textbf{wo HOR Sub-Net}& \textbf{Ours} \\ \hline\hline
\textbf{CSD(2000) PSNR/SSIM} & \textbf{32.94/0.94}   & \textbf{31.79/0.93}   & \textbf{29.76/0.93}      & \textbf{34.10/0.95}      \\ 

\bottomrule
\end{tabular}}
\label{tab:abalation4}
\end{table}

\subsubsection{Ablation study of the Efficient Pyramid Network.} To verify the effectiveness of each proposed component, we present comparison results in Table~\ref{tab:abalation4}. (a) \textbf{CIMB $\rightarrow$ SE-ResBlock}. We replace the proposed CIMB with SE-ResBlock~\cite{hu2018squeeze}. (b) \textbf{wo EAM}. We remove the proposed External Attention Module from our complete neural network. (d) \textbf{wo HOR Sub-set.} We remove the HOR Sub-net, only reserve our encoder and decoder. Each proposed component is necessary for our Efficient Pyramid Network.

\section{Limitations}
Compared with previous light-weight desnowing methods, for instance, HDCW-Net (only 699k Params.). The proposed Efficient Pyramid Network has better desnowing performance, but the much bigger parameters of our network will result in difficulty when deployed in edge devices.

\section{Conclusion}
In this work, we propose an Efficient Pyramid Network to handle High-Definition snow images in real time. Our extensive experiment and ablation study demonstrate the effectiveness of our proposed method and proposed blocks.

Although our method is simple, it is superior to all
the previous state-of-art desnowing methods with a considerable margin on three widely-used large-scale snow datasets. We hope to further promote our method to other
low-level vision tasks such as deraining and dehazing.



\bibliographystyle{splncs}
\bibliography{egbib}

\end{document}